\pdfoutput=1
\documentclass[11pt]{article}

\usepackage[]{EMNLP2023}

\usepackage{times}
\usepackage{latexsym}

\usepackage{amsmath}
\usepackage{amsthm}
\usepackage{booktabs}
\usepackage{multirow}
\usepackage{graphicx}
\usepackage{threeparttable}

\usepackage{enumitem}
\usepackage{float}
\usepackage{amsfonts}
\usepackage[ruled,vlined]{algorithm2e}
\usepackage{algorithmic}
\usepackage{url}

\usepackage[T1]{fontenc}
\usepackage[utf8]{inputenc}

\usepackage{microtype}

\usepackage{inconsolata}

\usepackage{graphicx}

\title{
Let the Pretrained Language Models "Imagine" for Short Texts Topic Modeling
}

\author{Pritom Saha Akash $\quad$ Jie Huang $\quad$ Kevin Chen-Chuan Chang \\
University of Illinois at Urbana-Champaign, USA \\
 \texttt{\{pakash2, jeffhj, kcchang\}@illinois.edu}
}

\begin{document}
\maketitle

\begin{abstract}
Topic models are one of the compelling methods for discovering latent semantics in a document collection. However, it assumes that a document has sufficient co-occurrence information to be effective. However, in short texts, co-occurrence information is minimal, which results in feature sparsity in document representation. Therefore, existing topic models (probabilistic or neural) mostly fail to mine patterns from them to generate coherent topics. In this paper, we take a new approach to short-text topic modeling to address the data-sparsity issue by extending short text into longer sequences using existing pre-trained language models (PLMs). Besides, we provide a simple solution extending a neural topic model to reduce the effect of noisy out-of-topics text generation from PLMs. We observe that our model can substantially improve the performance of short-text topic modeling. Extensive experiments on multiple real-world datasets under extreme data sparsity scenarios show that our models can generate high-quality topics outperforming state-of-the-art models. \footnote{Code and data will be released after the review process.}
\end{abstract}
\section{Introduction}
\label{sec:introduction}
In the digital era, short texts dominate the Web, such as tweets, web page titles, news headlines, image captions, product reviews, etc. These short texts are one of the most effective mediums for sharing knowledge. However, the volume of short texts is also huge because of the information explosion, which demands an external mechanism for extracting key information from them. Topic modeling is one such mechanism for uncovering latent topics from short texts, which has a wide range of applications, such as comment summarization \cite{ma2012topic}, content characterization \cite{ramage2010characterizing,zhao2011comparing}, emergent topic detection \cite{lin2010pet}, document classification \cite{sriram2010short}, user interest profiling \cite{weng2010twitterrank}, and so on.

Traditional topic models (e.g., LDA, PLSA) \cite{blei2003latent, hofmann1999probabilistic} are primarily used to discover latent topics from text corpora. However, these models largely assume that each given text document has rich context information to infer topic structures from the corpus. Therefore, the lack of ample context information in short texts makes topic modeling a challenging task. This issue is also called the data sparsity problem, where the co-occurrence information in short texts is minimal, making traditional models less effective in high-quality topic mining. 

There are several works for short-text topic modeling. One such simple but the popular strategy is to aggregate a subset of short texts into a longer pseudo document so that conventional topic models can be applied. This aggregation is guided by different metadata information. E.g., \citet{weng2010twitterrank} aggregated the tweets by the same user into a single document before applying LDA. Other metadata used for aggregation are hashtags \cite{mehrotra2013improving} and external corpora \cite{zuo2016topic} and so on. However, this metadata may not always be available. Therefore, another line of work uses inherent structural or semantic information, i.e., Biterm Topic Model (BTM) \cite{yan2013biterm} that infer topic distributions over unordered word pairs called biterms. GraphBTM extends this idea by extracting transitive features from biterms for creating topic models \cite{zhu2018graphbtm}. However, they are not generally able to generate the topic distribution for an individual document. Another strategy limits the number of active topics for each short text. E.g., \citet{yin2014dirichlet} sample each document from a single topic. However, this approach restricts a model's capacity because many short texts may cover more than one topic.

A short text (e.g., title, caption) is usually a summarized version of an existent longer text, providing an excellent hint to readers about the longer text. To judge the topics of a short text, humans usually ``\textit{imagine}'' the context of the short text. 
% E.g., from a news headline, "No tsunami but FIFA's corruption storm rages on",  people can easily understand this is about the topic "sports". Because in their imagination, they can gather context about "FIFA" and its relatedness to the sport of soccer.  
E.g., for a news headline: ``No tsunami but FIFA's corruption storm rages on'', humans may guess its content and gather context about ``FIFA'' through imagination; 
% gather context about ``FIFA'' and its relatedness to the sport of soccer through imagination; 
based on this, they can understand the headline is about the topic ``sports''. 

Now, can machines also ``\textit{imagine}'' the context to better understand the topics of a short text?
% A machine can not "imagine" the context by itself unless we explicitly provide the context.
Recently, large-scale pre-trained language models (PLMs) such as BART \cite{lewis2019bart}, T5 \cite{raffel2020exploring}, and GPT2 \cite{radford2019language} have appeared as amazing open-ended text generator capable of rendering surprisingly fluent text from a limited preceding context. 
% Previous studies have also shown the exciting benefits of generated text using PLMs in various applications such as machine translation \cite{conneau2019cross}, text summarization \cite{rothe2020leveraging}, and dialog systems \cite{zhang2019dialogpt}. 
% Therefore, the question is, can we use PLMs to generate text as context - an alternative to the "imagination" part of humans?
E.g., from the previously specified news headline, the PLM T5 generates an extended sequence (as shown in the second column of Table \ref{tab:error_example}) with tokens like ``Sepp Blatter'', ``Fernando Torres'', and ``kicking'' that are strongly related to sports soccer. Therefore, generating texts using PLMs conditioned on the short text seems intuitive (like human imagination) to enrich its context so that topic models can capture sufficient co-occurrence to infer meaningful topics. Here, the advantages are twofold. First, it tries to tackle the actual challenge of short text topic modeling by making the text large. Second, as PLMs are proven to generate fluent text conditioned on only minimal context, no extra information is required except the short text itself.

% Therefore, considering the potential of PLMs' text generation capability and the limitation of existing short-text topic models, in this paper, we explore PLMs for this task. 
Therefore, in this paper, we propose to leverage ``imagination'' of pre-trained language models for short-text topic modeling.
Specifically, we extend a short text into a long sequence using PLMs (e.g., BART \cite{lewis2019bart}, T5 \cite{raffel2020exploring} and GPT2 \cite{radford2019language}). And then, we use the extended text with existing topic models for inferring latent topics. The result shows promising improvement in topic quality over only using short texts. However, as PLMs-grounded generation does not use fine-tuning on the given task, it may generate coherent texts but with domain shift possibility. To handle this possible issue, we use extended text only as contextual information for a document and reconstruct the short text by adapting a neural topic model. Concretely, we extend Neural ProdLDA \cite{srivastava2017autoencoding} that uses a black-box variational inference \cite{ranganath2014black}, to incorporate contextualized representations from long texts and reconstruct the short texts in the decoder.  The proposed approaches consistently improve topic quality over existing general purpose and short-text topic modeling.  

To summarize, our \textbf{contributions} in this paper are the following. We are the first to explore PLMs-based text generation for short text topic modeling. We show that a simple approach that uses PLM-generated longer sequences with existing topic models provides improvement according to topic quality metrics. Second, to handle the domain shift problem, we design a solution by extending a neural VAE-based topic model. Finally, we conduct a comprehensive set of experiments on multiple datasets over different tasks, demonstrating our models' superiority against existing baselines.
\begin{figure*}[!tb]
\centering
\includegraphics[width=0.91\linewidth]{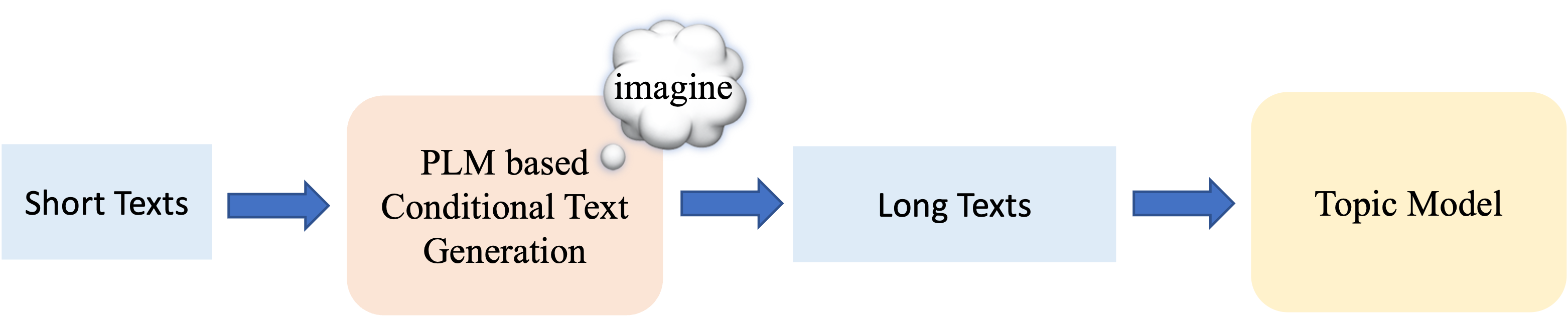}
\caption{Overview of the proposed architecture.}
\label{fig:arch0}
\vspace{-3mm}
\end{figure*}

\section{Proposed Methodology}
\label{sec:method}
Our proposed framework consists of two components. The first component generates longer text given a short text. The second one utilizes the generated longer texts for topic modeling. The overall framework is shown in Figure \ref{fig:arch0}.

\subsection{Short Text Extension}
\label{sec:short_exten}
We formulate the short text extension as a conditional sentence generation task, i.e., generating longer text sequences given a short text. Formally, we use the standard sequence-to-sequence generation formulation with a PLM $\mathcal{M}$: given input a short text sequence $x$, the probability of the generated long  sequence $y = [y_1, \dots, y_m]$ is calculated as:
\begin{align*}
    \mathbf{Pr}_{\mathcal{M}}(y|x) = \sum_{i=1}^m \mathbf{Pr}_{\mathcal{M}}(y_i|y_{<i}, x),
\end{align*}
where $y_{<i}$ denotes the previous tokens $y_1,\dots, y_{i-1}$. The PLM $\mathcal{M}$ specific text generation function $f_{\mathcal{M}}$ is used for sampling tokens and the sequence with the largest $ \mathbf{Pr}_{\mathcal{M}}(y|x)$ probability is chosen.

\subsection{Topic Model on Generated Long Text} 
\label{sec:LCSNTM}
Upon optioning the longer text sequences from the previous step, one possible straightforward way can be using existing topic models that work better for long text documents. As the longer texts have better co-occurrence context than the original short texts, it is expected to reduce the data sparsity problem of short-text topic modeling. Therefore, exploring existing probabilistic and neural topic models is intuitive on top of the generated longer text sequences. Therefore, we directly utilize different existing topic models on generated texts as one solution. 

\begin{figure}[!tb]
\centering
\includegraphics[width=0.9\linewidth]{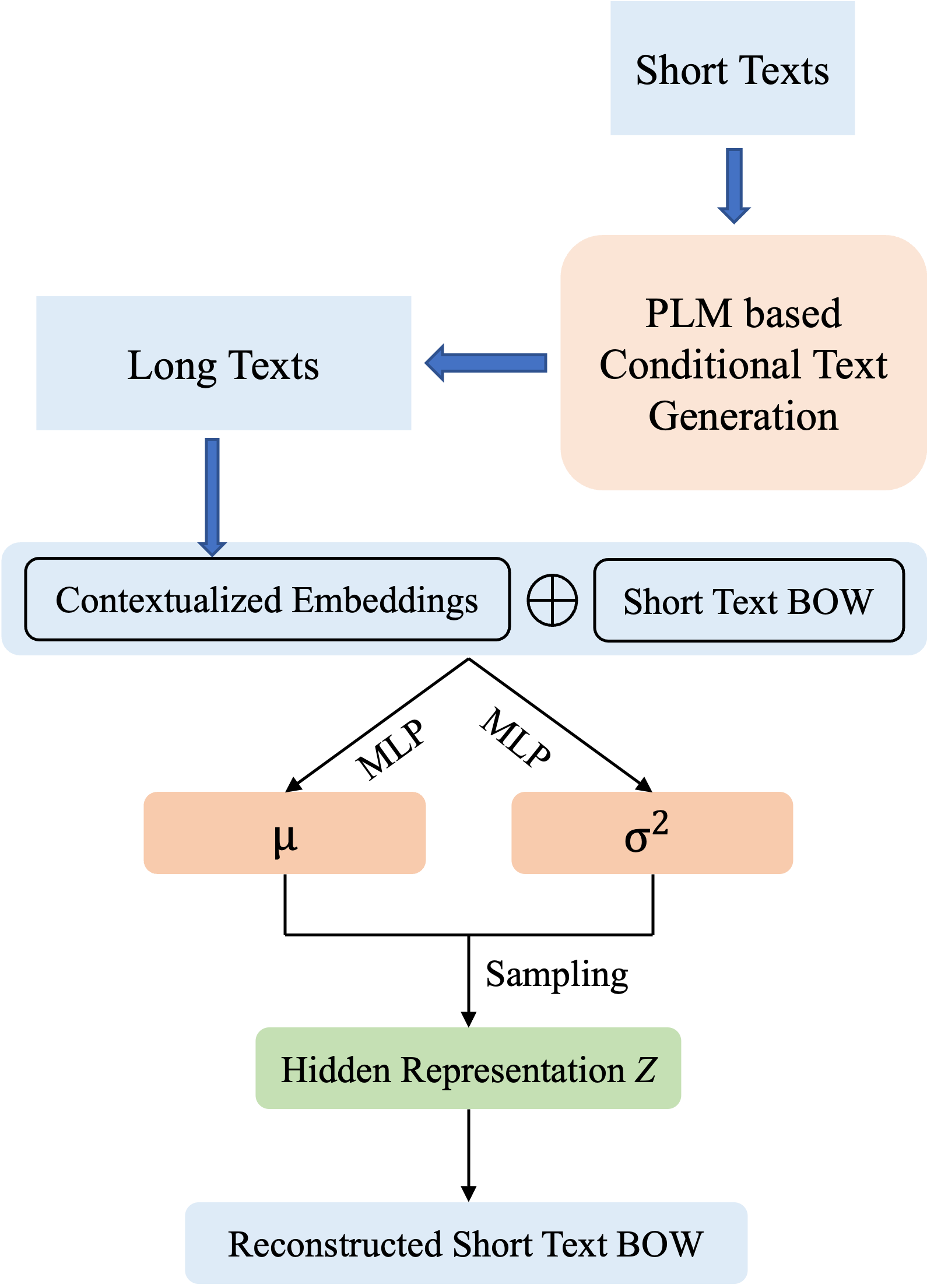}
\caption{Long Text Contextualized Short Text Neural Topic Model (LCSNTM).}
\label{fig:arch1}
\vspace{-5mm}
\end{figure}

However, as the pre-trained knowledge is directly used for text generation without finetuning on the target dataset, one possible issue with this straightforward approach is that the generated text may shift from the original domain or topic of the given short text (or partially cover the topics). One such inconsistency is shown in the third column of Table \ref{tab:error_example} where we see a longer sequence generated from a given short text using a PLM GPT-2. We observe that the generated sequence is coherent and easily readable sentences with many related words to the given short text. E.g., as the short text has content about the court proceeding, the generated long text has many such related words like ``judgment'', ``plaintiffs'' and so on. However, the generated text has partially shifted from the original topic of the text. More specifically, the "sports" aspect of the given short text is entirely missing in the generated longer text. Therefore, only relying on this generated text for topic modeling will likely miss the expected topics distribution in the result. To solve this issue, we propose a simple yet very effective solution by extending a neural topic model, which we call long text contextualized short text neural topic model (LCSNTM) as shown in Figure~\ref{fig:arch1}.

% \usepackage{booktabs}
% \begin{table*}[]
% \resizebox{1.0\linewidth}{!}{%
% \begin{tabular}{@{}ll@{}}
% \toprule
% Short Text                  & court agrees to expedite n.f.l.'s appeal                                                                                                \\ \midrule
% Generated Long Text (GPT2) & \begin{tabular}[c]{@{}l@{}}court agrees to expedite N.F.L. appeal.May 5, 1987 The Third United States Circuit Court of Appeals issues an \\order denying Enron's request for summary judgment in his suit seeking summary judgment from Enron in his\\ suit for injunctive relief to prevent Enron from misusing the trademark ""energy"" in commerce. Judge Joseph S.\\ Tumlinson's order states that both plaintiffs are estopped from arguing, pursuant to Fed.R.Civ.P....\end{tabular} \\ \bottomrule
% \end{tabular}}
% \caption{Example extended text generated using GPT2 from TagMyNews dataset.}
% \label{tab:error_example}
% \end{table*}

% \usepackage{booktabs}
\begin{table*}[!ht]
\resizebox{1.0\linewidth}{!}{%
\begin{tabular}{@{}lll@{}}
\toprule
\begin{tabular}[c]{@{}l@{}}Short \\ Texts\end{tabular}    & no tsunami but fifa's corruption storm rages on                                                                                                                                                                                                                                                                                                                                                                                                              & court agrees to expedite n.f.l.'s appeal                                                                                                                                                                                                                                                                                                                                                                                                     \\ \midrule
\begin{tabular}[c]{@{}l@{}}Extended \\ Texts\end{tabular} & \begin{tabular}[c]{@{}l@{}}no tsunami but fifa's corruption storm rages on. fifa president \\ sepp blatter speaks out about corruption scandals . but fifa's \\ stewardship is far from over and fifa are not at fault . Fernando \\ torres, fifa's head of integrity, is still alive and kicking . fa and \\ fifa must stop corruption before fifa takes over . fifa fans are not \\ safe when it comes to their vote, this is not the place..\end{tabular} & \begin{tabular}[c]{@{}l@{}}court agrees to expedite N.F.L. appeal.May 5, 1987. The Third \\ United States Circuit Court of Appeals issues an order denying\\ Enron's request for summary judgment in his suit seeking summary\\ judgment from Enron in his  suit for injunctive relief to prevent Enron \\ from misusing the trademark ""energy"" in commerce.\\ Judge Joseph S.Tumlinson's order states that both plaintiffs..\end{tabular} \\ \bottomrule
\end{tabular}}
\caption{Example short texts and corresponding extended texts using PLMs.}
\label{tab:error_example}
\end{table*}
{\flushleft \textbf{Long Text Contextualized Short Text Neural Topic Model:}}
As solely relying on generated long texts creates the problem of topic shift or incomplete topic coverage of a document, we use the generated sequence only as complementary information with given short text. Inspired by a previous work \cite{bianchi2020pre}, we incorporate the contextualized representation of generated long text along with the given short text bow as input of the topic model. This will enrich the context information of the given short text without much deviation from the original topics of the text. To further enforce this, we reconstruct the original short-text BOW rather than the generated long-text BOW. 

Formally, the model extends an existing topic model called ProdLDA \cite{srivastava2017autoencoding}. ProdLDA is a neural topic model based on the Variational AutoEncoder (VAE) mechanism \cite{kingma2013auto}. The encoder part of this model maps the BOW representation of a document to a continuous latent representation by training a neural variational inference network. More specifically, the model first generates mean vector $\mu$ and variance vector $\sigma^2$ by two separate MLPs from a document. The $\mu$ and $\sigma^2$ are then used to sample a latent representation $Z$ assuming Gaussian distribution. Then, a decoder network reconstructs the input BOW representation by generating its words from $Z$. In our model, instead of using only the short text BOW as input, we concatenate it with the contextualized representation of generated long text using an embedding representation (i.e., SBERT \cite{reimers2019sentence}). The model is trained with the original objective function \cite{srivastava2017autoencoding} called the evidence lower bound (ELBO) as follows:

\resizebox{0.9\linewidth}{!}{
\begin{minipage}{\linewidth}
\vspace{-4mm}
\begin{align}
    \mathcal{L}(\Theta) = \sum_{d\in \mathcal{D}} \sum_{n=1}^{N_d} \mathbb{E}_q[\log p(w_{dn} \mid Z_d)] - \nonumber \\
    \sum_{d\in \mathcal{D}} KL (q(Z_d;w_d, \Theta) \mid \mid p(Z_d)),
    \label{eq:elbo}
\end{align}
  \end{minipage}
}
where $w_{dn}$ is the $n$-th token in a document $d$ with length $N_d$ from the corpus $\mathcal{D}$. $\Theta$ represents learnable parameters in the model. $q(\cdot)$ is a Gaussian whose mean and variance are estimated from two separate MLPs.

\section{Experiments}
\label{sec:experiment}
In this section, we employ empirical evaluations, which are designed mainly to answer the following research questions (RQs):

\begin{itemize}[nolistsep,leftmargin=*]
 \item \textbf{RQ1.} Does the PLMs grounded text extension improves the performance of existing topic models over short texts in both cases of topic quality and text classification performance?
 
 \item \textbf{RQ2.} How effectively does the proposed LCSNTM improve the performance of topic modeling for short texts?
 
 \item \textbf{RQ3.} How qualitatively different are the topics discovered by the proposed architecture from existing baselines?

\end{itemize} 

\subsection{Experiment Setup}

{\flushleft \textbf{Datasets.}} We use the following datasets to evaluate our proposed architecture. The detailed statistics of these datasets are shown in Table \ref{tab:dataset_stat}.
\begin{itemize}[nolistsep,leftmargin=*]

    \item \textbf{StackOverflow:} 
     This dataset was created using the challenge information that was provided in 
     Kaggle\footnote{\url{https://www.kaggle.com/datasets/stackoverflow/stackoverflow}}. 
     We make use of the dataset that \citet{xu2015short} provided, which contains 20,000 randomly chosen question titles. Information technology terms like ``matlab'', ``osx'', and ``visual studio'' are labeled next to each question title.
    
    \item \textbf{TagMyNews:} Titles and contents of English news articles published by \citet{vitale2012classification} are included in this dataset . In our experiment, we use the headlines from the news as brief paragraphs. Every news item is given a ground-truth name, such as ``sci-tech'', ``business'', etc.
    
    \item \textbf{WebSnippets:} The web content from Google search snippets makes up the dataset provided by \citet{phan2008learning}. This dataset has eight labels, including ``Culture-Arts-Entertainment'' and ``Computers'' among others.
    
\end{itemize}

\begin{table}[]
\resizebox{1.0\linewidth}{!}{%
\begin{tabular}{@{}ccccc@{}}
\toprule
Datasets      & \# of docs & \begin{tabular}[c]{@{}c@{}}Average\\ length\end{tabular} & \begin{tabular}[c]{@{}c@{}}\# of class\\ labels\end{tabular} & \begin{tabular}[c]{@{}c@{}}Vocabulary\\ size\end{tabular} \\ \midrule
StackOverflow & 19899      & 4.49                                                     & 20                                                           & 2013                                                      \\
TagMyNews     & 4918       & 3.88                                                     & 7                                                            & 1410                                                      \\
WebSnippets   & 4067       & 14.52                                                    & 8                                                            & 12329                                                     \\ \bottomrule
\end{tabular}}
\caption{Statistics of datasets after preprocessing.}
\label{tab:dataset_stat}
\vspace{-1mm}
\end{table}

{\flushleft \textbf{Baselines.}}
We compare our models with the following baselines.
\begin{itemize}[nolistsep,leftmargin=*]
    \item \textbf{LDA:} We used one of the widely used probabilistic topic models, Latent Dirichlet Allocation (LDA) \cite{blei2003latent} as a baseline for this work.
    \item \textbf{CLNTM}: Contrastive Learning for Neural Topic Model combines contrastive learning paradigm with neural topic models by considering both effects of positive and negative pairs \cite{nguyen2021contrastive}.
    
    \item \textbf{CTM:} Contextualized Topic Model combines contextualized representations of documents with neural topic models \cite{bianchi2020pre}. 
    \item \textbf{BTM:} Biterm Topic Model \cite{yan2013biterm} uses extra structural information by directly constructing the topic distributions over unordered word pairs (biterms). This model is specialized for short text topic modeling.  
    \item \textbf{GraphBTM:} Another short text topic model called GraphBTM is an extension of BTM by extracting transitive features from biterms for creating topic models \cite{zhu2018graphbtm}.
    \item \textbf{NQTM:} NQTM is a neural topic model that employs a topic distribution quantization approach to generate peakier distributions that are better suited to modeling short texts \cite{wu2020short}.
\end{itemize}

{\flushleft \textbf{Pre-trained Language Models.}} We utilize three Pre-trained Language Models (PLMs): BART \cite{lewis2019bart}, T5 \cite{raffel2020exploring} and GPT-2 \cite{radford2019language}. They are separately used to conditionally generate longer text from a given short text.
\begin{itemize}[nolistsep,leftmargin=*]

 \item \textbf{BART\footnote{\href{https://huggingface.co/facebook/bart-large-cnn}{https://huggingface.co/facebook/bart-large-cnn}}:} We use the BART-Large-CNN, the large sized model pre-trained on English language and fine-tuned on CNN Daily Mail\footnote{\href{https://huggingface.co/datasets/cnn\_dailymail}{https://huggingface.co/datasets/cnn\_dailymail}}.

 \item \textbf{T5\footnote{\href{https://huggingface.co/t5-large}{https://huggingface.co/t5-large}}:} We use T5-Large model with the checkpoint of 770 million parameters. This model is pre-trained on the Colossal Clean Crawled Corpus (C4) \cite{raffel2020exploring}.
 
 \item \textbf{GPT-2\footnote{\href{https://huggingface.co/gpt2-large}{https://huggingface.co/gpt2-large}}:} We use GPT-2 Large model (774M parameter version of GPT-2), a transformer-based language model pretrained on English language using a causal language modeling (CLM) objective.
\end{itemize}
The implementation details are shown in Appendix~\ref{sec:imp_details}.

% {\flushleft \textbf{Implementation Details.}}
% There are some parameters for both the proposed architecture and baselines we need to set. For text generation from all three PLMs, we use the maximum new tokens length as 200 and the minimum length as 100. We find that using beam-search decoding with a beam size of 2 generates more coherent text for BART, while multinomial sampling works better in GPT-2 and T5 for all three datasets. The number of iterations for all the topic models is set to 100, except LDA uses 200 as the maximum number of iterations. For the contextualized representation of input documents in CTM and LCSNTM, we use pre-trained SBERT\footnote{\href{https://huggingface.co/sentence-transformers
% /paraphrase-distilroberta-base-v2}{https://huggingface.co/sentence-transformers/paraphrase-distilroberta-base-v2}} with a maximum sequence length of 512.
% All parameters during calculating evaluation metrics are set to the same value across all the models. E.g., the number of top words for each topic for calculating NPMI and IRBO is set to 10. In text classification experiments, we use the default parameters for MNB from scikit-learn\footnote{\href{https://scikit-learn.org}{https://scikit-learn.org}}. For SVM, we use the hinge loss with the maximum iteration of 5. For logistic regression, the maximum iteration is set to 1000, and the tree depth for RF is set to 3 with the number of trees as 200.

% \usepackage{booktabs}
\begin{table*}[!ht]
\centering
\resizebox{1.0\linewidth}{!}{%
\begin{tabular}{@{}cccc|cc|cc|cc|cc|cc|cc@{}}
\toprule
\multirow{3}{*}{Data}                & \multirow{3}{*}{\begin{tabular}[c]{@{}c@{}}Topic\\ Quality\\ Metrics\end{tabular}} & \multicolumn{6}{c}{General Purpose Topic Models}                              & \multicolumn{6}{c}{Short Text Topic Models}                                                                                  & \multicolumn{2}{c}{\multirow{3}{*}{LCSNTM}} \\ \cmidrule(lr){3-14}
                                     &                                                                                    & \multicolumn{2}{c}{LDA} & \multicolumn{2}{c}{CLNTM} & \multicolumn{2}{c}{CTM} & \multicolumn{2}{c}{BTM} & \multicolumn{2}{c}{\begin{tabular}[c]{@{}c@{}}Graph\\ BTM\end{tabular}} & \multicolumn{2}{c}{NQTM} & \multicolumn{2}{c}{}                      \\ \cmidrule(l){3-16} 
                                     &                                                                                    & k=20        & k=50      & k=20        & k=50        & k=20       & k=50       & k=20       & k=50       & k=20                               & k=50                               & k=20       & k=50        & k=20                & k=50                \\ \midrule
\multicolumn{16}{c}{StackOverflow}                                                                                                                                                                                                                                                                                                                                                                        \\ \midrule
\multirow{3}{*}{Short Text}               & {NPMI}                                                           & 0.041  & 0.011          & -0.074         & -0.141         & 0.021          & 0.099          & 0.062      & 0.074      & -0.154                             & -0.183                             & -0.08       & -0.099     & -                    & -                    \\
                                          & CWE                                                                                & 0.129  & 0.120          & 0.124          & 0.111          & 0.139          & 0.135          & 0.137      & 0.138      & 0.096                              & 0.097                              & 0.119       & 0.099      & -                    & -                    \\
                                          & IRBO                                                                                & 0.985  & 0.989          & 0.814          & 0.925          & 0.995          & 0.979          & 0.889      & 0.920      & 0.790                              & 0.952                              & 0.990       & 0.992      & -                    & -                    \\ \midrule
\multirow{3}{*}{Extended Text(BART)}      & {NPMI}                                                           & 0.042  & 0.014          & -0.054         & -0.051         & 0.034          & 0.056          & -          & -          & -                                  & -                                  & -           & -          & \textbf{0.101}       & \textbf{0.109}       \\ 
                                          & CWE                                                                                & 0.138  & 0.131          & 0.140          & 0.150          & \textbf{0.153}          & \textbf{0.155}          & -          & -          & -                                  & -                                  & -           & -          & 0.139                & 0.139                \\
                                          & IRBO                                                                                & 0.998  & \textbf{0.998} & \textbf{1.0}   & 0.996 & \textbf{1.0}   & 0.994          & -          & -          & -                                  & -                                  & -           & -          & 0.995                & 0.979                \\  \cmidrule(l){2-16}
\multirow{3}{*}{Extended Text(T5)}        & {NPMI}                                                           & 0.066  & 0.054          & 0.017          & 0.029          & 0.101          & \textbf{0.109} & -          & -          & -                                  & -                                  & -           & -          & \textbf{0.109}       & \textbf{0.109}       \\
                                          & CWE                                                                                & 0.139  & 0.145          & \textbf{0.174} & \textbf{0.167} & 0.139          & 0.157 & -          & -          & -                                  & -                                  & -           & -          & 0.137                & 0.140       \\
                                          & IRBO                                                                                & 0.998  & \textbf{0.997} & 0.997          & 0.996          & \textbf{1.0}   & 0.994          & -          & -          & -                                  & -                                  & -           & -          & 0.995                & 0.976                \\ \cmidrule(l){2-16}
\multirow{3}{*}{Extended Text(GPT-2)}      & {NPMI}                                                           & 0.08   & 0.089          & 0.005          & -0.007         & 0.094          & 0.098          & -          & -          & -                                  & -                                  & -           & -          & \textbf{0.102}       & \textbf{0.115}       \\
                                          & CWE                                                                                & 0.158  & 0.153          & \textbf{0.172}          & \textbf{0.178} & 0.148          & 0.158          & -          & -          & -                                  & -                                  & -           & -          & 0.145                & 0.140      \\
                                          & IRBO                                                                                & 0.990  & 0.992          & \textbf{0.999} & 0.996          & \textbf{0.996} & 0.995          & -          & -          & -                                  & -                                  & -           & -          & 0.988                & 0.967                \\ \midrule
\multicolumn{16}{c}{TagMyNews}                                                                                                                                                                                                                                                                                                                                                                            \\ \midrule
\multirow{3}{*}{Short Text}               & {NPMI}                                                           & -0.040 & -0.062         & -0.063         & -0.086         & -0.009         & -0.006         & -0.032     & -0.029     & -0.134                             & -0.140                             & -0.059      & -0.057     & -                    & -                    \\
                                          & CWE                                                                                & 0.107  & 0.096          & 0.104          & 0.088          & 0.158          & 0.164          & 0.127      & 0.124      & 0.072                              & 0.074                              & 0.095       & 0.092      & -                    & -                    \\
                                          & IRBO                                                                                & 0.998  & 0.999          & 0.752          & 0.978          & 0.996          & 0.982          & 0.971      & 0.975      & 0.960                              & 0.986                              & 0.959       & 0.951      & -                    & -                    \\ \midrule
\multirow{3}{*}{Extended Text(BART)}      & {NPMI}                                                           & 0.019  & 0.016          & 0.014          & 0.033          & \textbf{0.046} & \textbf{0.040} & -          & -          & -                                  & -                                  & -           & -          & 0.015                & 0.007                \\
                                          & CWE                                                                                & 0.171  & 0.163          & \textbf{0.247} & \textbf{0.208} & 0.199          & 0.197          & -          & -          & -                                  & -                                  & -           & -          & 0.168                & 0.165       \\
                                          & IRBO                                                                                & 0.983  & 0.993          & \textbf{1.000} & 0.996          & 0.999 & \textbf{0.995} & -          & -          & -                                  & -                                  & -           & -          & 0.992                & 0.983                \\
                                           \cmidrule(l){2-16}
\multirow{3}{*}{Extended Text(T5)}        & {NPMI}                                                           & -0.001 & -0.012         & -0.022         & 0.016          & \textbf{0.034} & \textbf{0.039} & -          & -          & -                                  & -                                  & -           & -          & 0.007                & 0.012                \\
                                          & CWE                                                                                & 0.156  & 0.153          & \textbf{0.252} & \textbf{0.215} & 0.211          & 0.201          & -          & -          & -                                  & -                                  & -           & -          & 0.161                & 0.153       \\
                                          & IRBO                                                                                & 0.964  & 0.990          & \textbf{1.000} & \textbf{0.997} & 0.999          & 0.994          & -          & -          & -                                  & -                                  & -           & -          & 0.993                & 0.978                \\ \cmidrule(l){2-16}
\multirow{3}{*}{Extended Text(GPT-2)}      & {NPMI}                                                           & 0.035  & 0.031          & 0.018          & \textbf{0.066} & \textbf{0.065} & 0.054          & -          & -          & -                                  & -                                  & -           & -          & 0.009                & 0.001                \\
                                          & CWE                                                                                & 0.198  & 0.185          & \textbf{0.273} & \textbf{0.253}          & 0.231          & 0.222          & -          & -          & -                                  & -                                  & -           & -          & 0.162                & 0.162                \\
                                          & IRBO                                                                                & 0.951  & 0.981          & \textbf{1.000} & \textbf{0.998} & 0.999          & 0.994          & -          & -          & -                                  & -                                  & -           & -          & 0.985                & 0.971                \\ \midrule
\multicolumn{16}{c}{WebSnippets}                                                                                                                                                                                                                                                                                                                                                                          \\ \midrule
\multirow{3}{*}{Short Text}               & {NPMI}                                                           & -0.045 & -0.061         & -0.110         & -0.059         & 0.002          & 0.001          & 0.009      & 0.01       & -0.154                             & -0.136                             & -0.177      & -0.156     & -                    & -                    \\
                                          & CWE                                                                                & 0.163  & 0.144          & 0.248          & 0.188          & 0.209          & 0.224          & 0.192      & 0.188      & 0.115                              & 0.107                              & 0.096       & 0.091      & -                    & -                    \\
                                          & IRBO                                                                                & 0.995  & 0.997          & \textbf{1.000}          & 0.759          & 0.999          & 0.998          & 0.918      & 0.953      & 0.944                              & 0.972                              & 0.996       & 0.992      & -                    & -                    \\ \midrule
\multirow{3}{*}{Extended Text(BART)}      & {NPMI}                                                           & -0.019 & -0.039         & -0.092         & -0.090         & \textbf{0.030} & 0.001          & -          & -          & -                                  & -                                  & -           & -          & 0.025                & \textbf{0.109}       \\
                                          & CWE                                                                                & 0.150  & 0.155          & 0.202          & 0.208          & 0.219          & 0.210          & -          & -          & -                                  & -                                  & -           & -          & \textbf{0.226}       & 0.220                \\
                                          & IRBO                                                                                & 0.996  & \textbf{0.999} & 0.999          & 0.998          & \textbf{1.000}          & 0.997          & -          & -          & -                                  & -                                  & -           & -          & \textbf{1.000}       & 0.996                \\ \cmidrule(l){2-16}
\multirow{3}{*}{Extended Text(T5)}        & {NPMI}                                                           & 0.008  & -0.035         & -0.095         & -0.074         & \textbf{0.033} & 0.013          & -          & -          & -                                  & -                                  & -           & -          & 0.012                & \textbf{0.115}       \\
                                          & CWE                                                                                & 0.193  & 0.170          & 0.230          & 0.230          & \textbf{0.246} & 0.234          & -          & -          & -                                  & -                                  & -           & -          & 0.234                & 0.221                \\
                                          & IRBO                                                                                & 0.996  & 0.998          & 0.999 & 0.998          & \textbf{1.000} & \textbf{1.000} & -          & -          & -                                  & -                                  & -           & -          & \textbf{1.000}       & 0.995                \\ \cmidrule(l){2-16}
\multirow{3}{*}{Extended Text(GPT-2)}      & {NPMI}                                                           & 0.013  & -0.024         & -0.048         & -0.058         & 0.020          & 0.008          & -          & -          & -                                  & -                                  & -           & -          & \textbf{0.028}       & \textbf{0.109}       \\
                                          & CWE                                                                                & 0.202  & 0.181          & 0.232          & \textbf{0.246} & \textbf{0.250} & 0.234          & -          & -          & -                                  & -                                  & -           & -          & 0.241                & 0.221                \\
                                          & IRBO                                                                                & 0.998  & \textbf{0.999} & 0.999          & 0.998          & \textbf{1.000} & 0.996          & -          & -          & -                                  & -                                  & -           & -          & \textbf{1.000}       & 0.993                \\  \bottomrule 
\end{tabular}
}
\caption{Topic coherences (NPMI and CWE) and diversity (IRBO) scores of topic words. $k$ is the topic number. The best in each case is shown in \textbf{bold}.}
\label{tab:topic_quality}
% \vspace{-1mm}
\end{table*}

\subsection{Topic Quality Evaluation} 

{\flushleft \textbf{Evaluation Metrics.}}
We evaluate each model using two different metrics: two for topic coherence (i.e., NPMI and CWE) and one for topic solution diversity (i.e., IRBO).
\begin{itemize}[nolistsep,leftmargin=*]

 \item \textbf{NPMI}: Normalized Pointwise Mutual Information (NPMI) is a standard measure of interpretability based on the average pointwise mutual information between randomly drawn two words from the same document \cite{lau2014machine}.

\item \textbf{CWE:} Coherence Word Embeddings (CWE) \cite{fang2016using} metric uses semantic similarity by word embeddings for calculating coherence in a topic model. As NPMI looks for actual co-occurrence between words, it may lose the semantic relatedness while calculating the coherency. In such cases, CWE is complementary to provide a complete view of the coherency of a topic model.

 \item \textbf{IRBO:} Inverted Rank-Biased Overlap (IRBO) evaluates the topic diversity by calculating rank-biased overlap over the generated topics introduced in \cite{webber2010similarity}. 
\end{itemize}

{\flushleft \textbf{Results and Discussions.}}
We first analyze the result of existing topic models on the generated text from PLMs (described in Section \ref{sec:short_exten}). The topic quality scores (NPMI, CWE, and IRBO) in Table~\ref{tab:topic_quality} show the apparent dominance of topic models on extended text compared to short texts. The best NPMI and IRBO scores for all three datasets are from either three extended texts  (i.e., BART, T5, or GPT-2) with significant improvement in topic coherency and comparable diversity. This clearly shows that extension of short text using PLMs help discover higher-quality topics that are more coherent and diverse. E.g., in CLNTM, the coherence score CWE gets improved $\sim$162\% (similarly $\sim$130\% in NPMI) from when using short text to GPT-2 generated extended sequence. However, these topic quality results do not always show that the mined topics correctly represent the target dataset. As specified in Section \ref{sec:LCSNTM}, the topics may shift because of the PLM-generated texts. We further discuss this through classification results in the next section.

Now, considering the topic quality performance of the proposed LCSNTM, we find some interesting findings. In almost all cases, we get an improvement in topic quality scores compared to the short-text counterparts. More specifically, in Stackoverflow and WebSnippets datasets, we obtained a significant performance boost in terms of NPMI coherence score compared to all other baselines with comparable CWE and IRBO scores. E.g., in the WebSnippets dataset, compared to the most similar model CTM,  the NPMI score for LCSNTM increases from 0.001 to 0.115 (with a 114\% improvement). 

However, in the TagMyNews dataset, the improvement in topic quality is not as promising as baselines on extended texts. One possible reason for this is that this dataset's average document text length is extremely short (i.e., as shown in Table~\ref{tab:dataset_stat}). And each of these short texts carries very limited (or absent) topic-indicative words. Therefore, while the LCSNTM reconstructs this short text during training, the generated topics may become less coherent. On the other hand, for the baselines that solely use the generated long texts, this problem is resolved by coherent tokens from the extended texts.

\subsection{Text Classification Evaluation}
Although text classification is not the main purpose of topic models, the generated document topic distribution can be used as the document feature for learning text classifiers. Therefore, we evaluate how learned document topic distribution is distinctive and informative enough to represent a document to be used for classifying a document correctly. We employ four different classification models on top of document topic distribution learned by different models. The classification models are Multinomial Naive Bayes classifier (MNB) \cite{kibriya2004multinomial}, Support Vector Machine (SVM) \cite{cortes1995support}, Logistic Regression (LR) \cite{wright1995logistic}, and Random Forest (RF) \cite{breiman2001random}. We use classification accuracy over 5-fold cross-validation to compare the performance of multiple classifiers. As BTM and GraphBTM are not designed to generate document-level topic distribution, we exclude these two in the text classification experiment. 

% \usepackage{booktabs}
\begin{table*}[!ht]
\resizebox{1.0\linewidth}{!}{%
\begin{tabular}{@{}ccccc|cccc|cccc|cccc|cccc@{}}
\toprule
                    & \multicolumn{4}{c}{LDA}                                                                                       & \multicolumn{4}{c}{CLNTM}                                                                                     & \multicolumn{4}{c}{CTM}                                                                                       & \multicolumn{4}{c}{NQTM}                                                                                      & \multicolumn{4}{c}{LCSNTM}                                                                                                                  \\ \midrule
                    &{MNB}   &{SVM}   &{LR}    &{RF}    &{MNB}   &{SVM}   &{LR}    &{RF}    &{MNB}   &{SVM}   &{LR}    &{RF}    & MNB                       & SVM                       & LR                        & RF                        & MNB                                & SVM                                & LR                        & RF                                 \\
                    \midrule
\multicolumn{21}{c}{StackOverflow}                                                                                                                                                                                                                                                                                                                                                                                                                                                              \\ \midrule
Short Text          &{0.643} &{0.617} &{0.643} &{0.586} &{0.051} &{0.051} &{0.066} &{0.095} &{0.807} &{\textbf{0.832}} &{0.833} &{0.770} & 0.050                     & 0.050                     & 0.050                     & 0.050                     & -                                  & -                                  & -                         & -                                  \\

Extended Text(BART) &{0.561} &{0.567} &{0.598} &{0.546} &{0.603} &{0.546} &{0.668} &{0.541} &{0.613} &{0.680} &{0.680} &{0.648} & -                         & -                         & -                         & -                         & 0.812                              & 0.824                              & 0.833                     & \textbf{0.775}                     \\

Extended Text(T5)   &{0.605} &{0.584} &{0.618} &{0.556} &{0.594} &{0.517} &{0.656} &{0.501} &{0.658} &{0.693} &{0.710} &{0.637} & -                         & -                         & -                         & -                         & \textbf{0.815}                     & 0.829                     & \textbf{0.834}            & 0.739                              \\

Extended Text(GPT-2) &{0.557} &{0.548} &{0.583} &{0.515} &{0.561} &{0.522} &{0.604} &{0.486} &{0.572} &{0.587} &{0.613} &{0.544} & -                         & -                         & -                         & -                         & 0.795                              & 0.796                              & 0.803                     & 0.695                              \\

 \midrule
\multicolumn{21}{c}{TagMyNews}                                                                                                                                                                                                                                                                                                                                                                                                                                                                                                                                                                                                 \\ \midrule
Short Text          & 0.335                     & 0.328                     & 0.370                     & 0.311                     & 0.254                     & 0.187                     & 0.264                     & 0.262                     & 0.564                     & 0.662                     & 0.675                     & 0.529                     & {0.274} & {0.143} & {0.278} & {0.282} & -                                  & -                                  & -                         & -                                  \\

Extended Text(BART) & 0.548                     & 0.588                     & 0.611                     & 0.491                     & 0.600                     & 0.628                     & 0.633                     & 0.470                     & 0.540                     & 0.664                     & 0.674                     & 0.524                     & -                         & -                         & -                         & -                         & {0.570}          & {0.672}          & {0.682} & {0.531}          \\

Extended Text(T5)   & 0.564                     & 0.599                     & 0.631                     & 0.477                     & \textbf{0.660}            & 0.662                     & 0.676                     & 0.509                     & 0.614                     & \textbf{0.717}            & \textbf{0.732}            & 0.557                     & -                         & -                         & -                         & -                         & {0.565}          & {0.671}          & {0.676} & {\textbf{0.557}} \\

Extended Text(GPT-2) & 0.550                     & 0.604                     & 0.617                     & 0.470                     & 0.611                     & 0.606                     & 0.624                     & 0.470                     & 0.505                     & 0.634                     & 0.6386                    & 0.489                     & -                         & -                         & -                         & -                         & {0.565}          & {0.650}          & {0.657} & {0.501}          \\

\midrule
\multicolumn{21}{c}{WebSnippets}                                                                                                                                                                                                                                                                                                                                                                                                                                                                                                                                                                                               \\ \midrule
Short Text          & 0.531                     & 0.575                     & 0.591                     & 0.402                     & 0.215                     & 0.150                     & 0.472                     & \textbf{0.716}                     & 0.712                     & 0.850                     & \textbf{0.856}            & 0.632                     & {0.397} & {0.380} & {0.438} & {0.376} & -                                  & -                                  & -                         & -                                  \\

Extended Text(BART) & 0.547                     & 0.621                     & 0.628                     & 0.486                     & 0.653                     & 0.765                     & 0.773                     & 0.603                     & 0.589                     & 0.792                     & 0.799                     & 0.628                     & -                         & -                         & -                         & -                         & {\textbf{0.720}} & {0.829}          & {0.850} & {0.678} \\

Extended Text(T5)   & 0.657                     & 0.717                     & 0.724                     & 0.532                     & 0.712                     & 0.801                     & 0.820                     & 0.648                     & 0.696                     & 0.826                     & 0.843                     & 0.601                     & -                         & -                         & -                         & -                         & {0.703}          & {\textbf{0.852}} & {0.852} & {0.647}          \\

Extended Text(GPT-2) & 0.564                     & 0.637                     & 0.640                     & 0.532                     & 0.529                     & 0.695                     & 0.696                     & 0.607                     & 0.495                     & 0.701                     & 0.690                     & 0.546                     & -                         & -                         & -                         & -                         & {0.654}          & {0.817}          & {0.820} & {0.682}          \\

 \bottomrule
\end{tabular}}
\caption{Text classification accuracy over 5-fold cross validation. The best results in each case are shown in \textbf{bold}.}
\label{tab:classification}
\vspace{-1mm}
\end{table*}

{\flushleft \textbf{Results and Discussions.}}
The classification result is presented in Table \ref{tab:classification}. Overall, the proposed LCSNTM is the best-performing model regarding classification accuracy, leveraging both the generated text and considering the topics shift (or incomplete coverage of topics) problem. As specified before, when using PLMs without finetuning on the target corpus, the generated text may not cover the original topics of the document or shift from them. This issue is also visible in the classification result among baselines that directly use the generated longer text for topic modeling. E.g., we can see a substantial performance drop in accuracy in the StackOverflow dataset (e.g., from 0.807 to 0.572 in MNB while using CTM with GPT-2-generated text). Even if the StackOverflow dataset is about a particular technical domain, the PLMs are more likely to generate tokens from general domains. That is why the learned topics from the extended texts may not represent the original documents, resulting in poor classification performance.
This effect is comparatively less in the other two datasets, as those are about more general topics like ``politics'', ``sports'', etc. On the other hand, the LCSNTM reduces this effect by reconstructing the original short texts during training which is also visible in the classification result.  

From the above results, it is evident that LCSNTM makes a tradeoff between topic quality and classification performance, while others improve in one direction only. 

We have also shown the effect of the different generated text sizes on the topic quality in Appendix \ref{sec:text_length}.

% \usepackage{booktabs}
\begin{table}[tp!]
\centering
\resizebox{1.0\linewidth}{!}{%
\begin{tabular}{@{}p{0.19\linewidth}l|l@{}}
% \begin{tabular}{@{}ll|l@{}}
\toprule
Models   & \begin{tabular}[c]{@{}l@{}}Topic Words\\ (on Short Text)\end{tabular}                                                                                         & \begin{tabular}[c]{@{}l@{}}Topic Words\\ (on GPT-2 Long Text)\end{tabular}                                                                    
                                            \\ \midrule
LDA      & \begin{tabular}[c]{@{}l@{}}application window load open test\\ linq oracle sql query table\\ matlab update image value field\end{tabular}                     & \begin{tabular}[c]{@{}l@{}}application spring api java library\\ database query table sql oracle\\ matlab image number size color\end{tabular}      \\ \midrule
CLNTM    & \begin{tabular}[c]{@{}l@{}}pl sql outer procedure join\\ pl sql script mark os\\ script pl sql linqtosql not\end{tabular}                                     & \begin{tabular}[c]{@{}l@{}}clause join query hql desc\\ ipad usb iphone icloud player\\ maven tomcat npm gradle restful\end{tabular}                \\ \midrule
CTM      & \begin{tabular}[c]{@{}l@{}}good best framework way web\\ scala class method java object\\ mac os osx run application\end{tabular}                             & \begin{tabular}[c]{@{}l@{}}bash script shell command path\\ svn repository git subversion branch\\ sql database query oracle statement\end{tabular} \\ \midrule
BTM      & \begin{tabular}[c]{@{}l@{}}use file visual excel studio\\ use file magento drupal hibernate\\ use magento file oracle way\end{tabular}                        & \multicolumn{1}{c}{-}                                                                                                                               \\ \midrule
GraphBTM & \begin{tabular}[c]{@{}l@{}}example axis applescript log properly\\ derive hold partition line spreadsheet\\ applescript parent hold example axis\end{tabular} & \multicolumn{1}{c}{-}                                                                                                                               \\ \midrule
NQTM     & \begin{tabular}[c]{@{}l@{}}custom bit lambda depth map\\ specific crash dead svn handling\\ use file excel wordpress magento\end{tabular}                     & \multicolumn{1}{c}{-}                                                                                                                               \\ \midrule
    LCSNTM      & \multicolumn{1}{c|}{-}                                                                                                                                         & \begin{tabular}[c]{@{}l@{}}oracle database sql store procedure\\ bash script command line shell \\ ajax apache request rewrite jquery\end{tabular}  \\ \bottomrule
\end{tabular}}
\caption{Topic words examples under k = 20.}
\label{tab:qualitative}
\vspace{-2mm}
\end{table}

\subsection{Topic Examples Evaluation}
To evaluate the proposed models qualitatively, we show the top five words for each of the three topics generated by different models in Table \ref{tab:qualitative}. We observe that some models on short texts generate topics with repetitive words (e.g., CLNTMa and BTM). Although the CTM on short texts generates diverse topics, they are less informative (i.e., with words like ``best'', ``good'', etc.). On the other hand, topics in generated long texts are less repetitive with much more coherency, although some also tend to generate topics with general words like ``number'' and ``size''. Finally, the LCSNTM generates both non-repetitive and informative topics. E.g., it is easy to detect that the three discovered topics are database, shell, and web programming.

\section{Related Work}
\label{sec:related_work}

\subsection{Traditional Topic Models}
The widely used traditional topic models, also known as probabilistic topic models, such as Probabilistic Latent Semantic Analysis (PLSA) \cite{hofmann1999probabilistic} and Latent Dirichlet Allocation (LDA) \cite{blei2003latent}, performs well when the given corpus consists of large-sized documents. These models assume that the documents have sufficient co-occurrence information to capture latent topic structures from the corpus. Thus, these models typically fail to infer high-quality topics from short texts such as news titles and image captions. To solve this issue, one strategy in existing probabilistic topic models uses structural and semantic information from texts such as Biterm Topic Model (BTM) \cite{yan2013biterm}. Another strategy aggregates a subset of short texts into a longer pseudo document using various metadata (e.g., hashtags, external corpora) before applying conventional topic models \cite{mehrotra2013improving,zuo2016topic}.  
Another line of short-text topic modeling restricts the document-topic distribution by assuming each document is sampled from a single topic such as Dirichlet Multinomial Mixture (DMM) model \cite{yin2014dirichlet,nigam2000text}. Although this is intuitive considering the limited context in shorts, this simplification may be too strict in practice as many short texts could cover more than one topic.

\subsection{Neural Topic Models}

With the recent developments in deep neural networks (DNNs) and deep generative models, there has been an active research direction in leveraging DNNs for inferring topics from corpus, also called neural topic modeling. The recent success of variational autoencoders (VAE) \cite{kingma2013auto} has opened a new research direction for neural topic modeling \cite{nan2019topic}. The first work that uses VAE for topic modeling is called the Neural Variational Document Model (NVDM) \cite{miao2016neural}, which leverages the reparameterization trick of Gaussian distributions and achieves a fantastic performance boost. Another related work called ProdLDA \cite{srivastava2017autoencoding} uses Logistic Normal distribution to handle the difficulty of the reparameterization trick for Dirichlet distribution. 

There also have been several works in neural topic modeling (NTM) for short texts. E.g., \cite{zeng2018topic} combines NTM with a memory network for short text classification. \cite{zhu-etal-2018-graphbtm} takes the idea of the probabilistic biterm topic model to NTM where the encoder is a graph neural network (GNN) of sampled biterms. However, this model is not generally able to generate the topic distribution of an individual document. \cite{lin2020copula} introduce the Archimedean copulas idea in the neural topic model to regularise the discreteness of topic distributions for short texts, which restricts the document from some salient topics. From a similar intuition, \cite{feng2022context} proposes an NTM by limiting the number of active topics for each short document and also incorporating the word distributions of the topics from pre-trained word embeddings. Another neural topic model \cite{wu2020short} employs a topic distribution quantization approach to generate peakier distributions that are better suited to modeling short texts.

\subsection{PLMs in Topic Models}
Previously, some neural topic models attempted to use PLMs as input representations of given documents.
E.g., a model called the contextualized topic model (CTM) \cite{bianchi2020pre} complements the Bag of Words (BOW) representation of a document with its contextualized vector representation from PLMs like BERT \cite{devlin2018bert}. As PLMs are pre-trained on large-scale text corpora such as Wikipedia and hold rich linguistic features, they are supposed to capture the context and order information in a text ignored in BOW representation. Similarly, BERTopic \cite{grootendorst2022bertopic} also uses PLM-based document embedding to cluster them and TF-IDF to find representative words from each cluster as topics. However, as it uses TF-IDF metrics, it fails to take benefit of the distributed representations of PLMs, which are better at capturing word semantics than frequency-based statistics.
Moreover, the above approaches do not solve the data sparsity problem in short text topic modeling but rather use PLMs only for better representation of input documents for general-purpose topic modeling. Unlike these neural topic models, the proposed framework in this paper uses PLMs to enrich contextual information of short documents by conditional text generation.
\section{Conclusion}
In this paper, we proposed a simple yet effective approach for short-text topic modeling leveraging the ``imagination'' capability of PLMs. To solve the data-sparsity problem of short texts, we first extend them into longer sequences using a PLM. These longer sequences are then used to mine topics by existing topic models. To further reduce the effect of the domain-shift problem of a pre-trained model, we propose a solution extending a neural topic model. A set of empirical evaluations demonstrate the effectiveness of the proposed framework over the state-of-the-art.

\section*{Limitations}
The proposed framework directly utilize PLMs for text generation conditioned on the given short texts. As we have specified before, this may result in noisy out-of-domain text generation, which hurts the document representativeness of the generated topics. This problem may worsen when the target domain is very specific. Although the proposed LCSNTM tries to solve this problem by a simple mechanism of short text reconstruction, it does not work in extreme sparsity scenarios, as we observed in the TagMyNews dataset. Therefore, controlling the generation process such that it outputs more relevant text in the target domain is a possible future research direction in this line.
\bibliography{anthology}
\bibliographystyle{acl_natbib}

\clearpage
\appendix
\section{Implementation Details.}
\label{sec:imp_details}
% {\flushleft \textbf{Implementation Details.}}
There are some parameters for both the proposed architecture and baselines we need to set. For text generation from all three PLMs, we use the maximum new tokens length as 200 and the minimum length as 100. We find that using beam-search decoding with a beam size of 2 generates more coherent text for BART, while multinomial sampling works better in GPT-2 and T5 for all three datasets. The number of iterations for all the topic models is set to 100, except LDA uses 200 as the maximum number of iterations. For the contextualized representation of input documents in CTM and LCSNTM, we use pre-trained SBERT\footnote{\href{https://huggingface.co/sentence-transformers
/paraphrase-distilroberta-base-v2}{https://huggingface.co/sentence-transformers/paraphrase-distilroberta-base-v2}} with a maximum sequence length of 512.
All parameters during calculating evaluation metrics are set to the same value across all the models. E.g., the number of top words for each topic for calculating NPMI and IRBO is set to 10. In text classification experiments, we use the default parameters for MNB from scikit-learn\footnote{\href{https://scikit-learn.org}{https://scikit-learn.org}}. For SVM, we use the hinge loss with the maximum iteration of 5. For logistic regression, the maximum iteration is set to 1000, and the tree depth for RF is set to 3 with the number of trees as 200.

\section{Effect of extended text lengths}
\label{sec:text_length}

In this section, we analyzed the effect of generated text length on the topic quality (shown in \ref{tab:length}). Here, we use GPT2 on CTM (as it purely uses extended texts, the effects will be easily analyzed). We use different generated text sizes of 10, 20, 50, and 100.
Here, for almost all the cases, we can see improvement in topic quality in coherence (NPMI, CWE) when we increase the minimum generated sequence length with stable diversity scores (IRBO). This shows that when we have more context in the generated text, the learned topics are more coherent (interpretable) without hampering diversity. 

% \usepackage{booktabs}
\begin{table}[]
\resizebox{1.0\linewidth}{!}{%
\begin{tabular}{@{}l|l|l|l|l@{}}
\toprule
\multicolumn{1}{c}{Text-Length} & 20     & 30     & 50     & 100   \\ \midrule
\multicolumn{5}{c}{Stack Overflow}                                 \\ \midrule
NPMI                            & 0.072  & 0.077  & 0.082  & 0.083 \\
CWE                             & 0.157  & 0.158  & 0.159  & 0.153 \\
IRBO                            & 0.992  & 0.992  & 0.992  & 0.994 \\ \midrule
\multicolumn{5}{c}{TagMyNews}                                      \\ \midrule
NPMI                            & 0.032  & 0.037  & 0.044  & 0.045 \\
CWE                             & 0.189  & 0.201  & 0.199  & 0.201 \\
IRBO                            & 0.991  & 0.992  & 0.992  & 0.990 \\ \midrule
\multicolumn{5}{c}{WebSnippets}                                    \\ \midrule
NPMI                            & -0.015 & -0.028 & -0.008 & 0.008 \\
CWE                             & 0.227  & 0.212  & 0.237  & 0.234 \\
IRBO                            & 0.992  & 0.990  & 0.992  & 0.996 \\ \bottomrule
\end{tabular}}
\caption{Effect of generated text length on Topic quality}
\label{tab:length}
\end{table}

\end{document}